\documentclass[lettersize,journal]{IEEEtran}
\usepackage{epsfig,rotating,setspace,latexsym,amsmath,epsf,amssymb,amsfonts,bm,theorem,subfigure,epstopdf,graphicx}
\usepackage{cite,booktabs,authblk,tikz}

\usepackage{color}

\definecolor{commentcolor}{RGB}{110,154,155}   

\IEEEoverridecommandlockouts
\allowdisplaybreaks

\begin{document}
	\bstctlcite{IEEEexample:BSTcontrol} 
	
	
	
	\title{Augmenting Training Data with Vector-Quantized Variational Autoencoder for Classifying RF Signals} 
	
	\thispagestyle{plain}
	\pagestyle{plain}

	\author{Srihari Kamesh Kompella}
	\author[2]{Kemal Davaslioglu}
	\author[2]{Yalin E. Sagduyu}
	\author[2]{Sastry Kompella}
	\affil{\normalsize University of Illinois, Urbana-Champaign, IL, USA} 
	\affil[2]{\normalsize Nexcepta, Gaithersburg, MD, USA
		\thanks{This material is based upon work supported
by the ASA(ALT) SBIR CCoE under Contract No. W51701-24-C-0096.}
	}
	
	\maketitle
	
\begin{abstract}

Radio frequency (RF) communication has been an important part of civil and military communication for decades. With the increasing complexity of wireless environments and the growing number of devices sharing the spectrum, it has become critical to efficiently manage and classify the signals that populate these frequencies. In such scenarios, the accurate classification of wireless signals is essential for effective spectrum management, signal interception, and interference mitigation. However, the classification of wireless RF signals often faces challenges due to the limited availability of labeled training data, especially under low signal-to-noise ratio (SNR) conditions. To address these challenges, this paper proposes the use of a Vector-Quantized Variational Autoencoder (VQ-VAE) to augment training data, thereby enhancing the performance of a baseline wireless classifier. The VQ-VAE model generates high-fidelity synthetic RF signals, increasing the diversity and fidelity of the training dataset by capturing the complex variations inherent in RF communication signals. Our experimental results show that incorporating VQ-VAE-generated data significantly improves the classification accuracy of the baseline model, particularly in low SNR conditions. This augmentation leads to better generalization and robustness of the classifier, overcoming the constraints imposed by limited real-world data. By improving RF signal classification, the proposed approach enhances the efficacy of wireless communication in both civil and tactical settings, ensuring reliable and secure operations. This advancement supports critical decision-making and operational readiness in environments where communication fidelity is essential.				
\end{abstract}

	
\section{Introduction}\label{sec:Introduction}
	%
	%
	

Effective RF waveform classification is critical for emerging spectrum operations to allow for the identification of friendly and adversarial signals, spectrum management, the mitigation of jamming or interference, and dynamic spectrum sharing. However, a significant challenge in developing robust RF signal classification systems is the scarcity of labeled training data. This scarcity arises from the difficulty in collecting extensive and representative RF signal datasets, particularly under diverse operational conditions and low signal-to-noise ratio (SNR) environments \cite{Oshea3, shi2019deep}.
	
To overcome the limitations of traditional signal classification methods and the constraints of limited training data, deep learning approaches have emerged as promising solutions. Convolutional neural networks (CNNs), such as ResNet, have demonstrated the ability to learn complex patterns from wireless signal data, leading to improved classification accuracy. Nonetheless, these models typically require large volumes of labeled data to perform effectively, a requirement that is challenging to meet in the wireless domain. To address this challenge, data augmentation has been studied for wireless communications \cite{davaslioglu2018generative, erpek2018deep, tang2018digital, huang2019data, shi2022sensing, soltani2020more, ayanoglu2022machine}.
		
In this study, we tackle the RF waveform classification problem by developing a novel methodology that employs Vector-Quantized Variational Autoencoders (VQ-VAE) models \cite{van2017neural, razavi2019generating}. VQ-VAE models generate synthetic data that can accurately learn the wireless signal distribution and subtle variations in the data, thereby expanding the available dataset and improving the performance of deep learning classifiers. Trained VQ-VAE models are employed to produce synthetic training data. We have evaluated different sampling methods for the VQ-VAE such as posterior sampling, smapling from class centers, and interpolation between points. We have observed that most of these do not provide the required diversity to enrich the samples that are used in the classifier training. We develop a novel methodology where we inject noise in the latent space of the VQ-VAE model. We integrate these augmention during the training process of a ResNet-based classifier. Our empirical results using high frequency (HF) communication data show a substantial enhancement in classification accuracy using the VQ-VAE generated data. Specifically, our best model achieves an overall improvement of 4.06\% in classification accuracy over the baseline model that is trained only on the real data. Notably, in challenging low SNR conditions, such as -10~dB SNR, the model improves classification accuracy by 15.86\% compared to the baseline model.
	
Our contributions are twofold:
\begin{enumerate}
\item A novel framework to employ VQ-VAE models for wireless signal augmentation: We introduce a novel method to sample data from a VQ-VAE model to increase the data diversity and generate synthetic wireless signals, which addresses the challenge of limited training data in the wireless domain.
\item Enhanced classification performance at low SNR: We demonstrate significant improvements in wireless waveform classification accuracy, particularly in low SNR environments that enhances the robustness and reliability of signal classifiers.
\end{enumerate}
	
The remainder of this paper is organized as follows. Section~\ref{sec:system_model} describes wireless signal classification problem and presents a set of transformations tailored to augment RF waveforms during the training of the classification model. Section~\ref{sec:proposed_synth} discusses the proposed VQ-VAE-based synthetic data generation methodology. 
Section~\ref{sec:performance} presents our performance evaluations and discusses these results. In Section~\ref{sec:Conclusion}, we summarize our results and present future directions. 
	
\section{Wireless Signal Classification Problem}\label{sec:system_model}
For wireless signal classification problem, we focus on HF communication data, although our proposed approach is applicable to other wireless signal classification problems. HF communication is crucial for beyond line-of-sight (BLOS) scenarios, where direct line-of-sight links between transmitter and receiver are not feasible. This is achieved through ionospheric propagation, allowing signals to travel great distances by bouncing between the Earth's surface and the ionosphere \cite{whitehead2022novel, erpek2018routing, li2019joint, arikan2020receiver, hervas2020ionospheric}. This capability makes HF communication indispensable for various applications such as military communications, emergency response, and maritime communication, where the ability to maintain connectivity across rugged terrains and long distances is critical.


We consider 18 transmission modes commonly found in the HF bands \cite{scholl2019classification}. These modes include the following: MorseCode, PSK31,  PSK63, QPSK31, RTTY45/170, RTTY50/170, RTTY100/850, Olivia8/250, Olivia16/500, Olivia16/1000, Olivia32/1000, DominoEx, MT63\_1000, Navtex/Sitor-B, Single-Sideband upper (USB), Single-Sideband lower, AM broadcast, and HF/radiofax. Table~\ref{table:mods} summarizes the modulation and Baud rates of these waveforms. 

\begin{figure}[tb!]
        \centerline{\includegraphics[width=0.8\linewidth,height=2.25in]{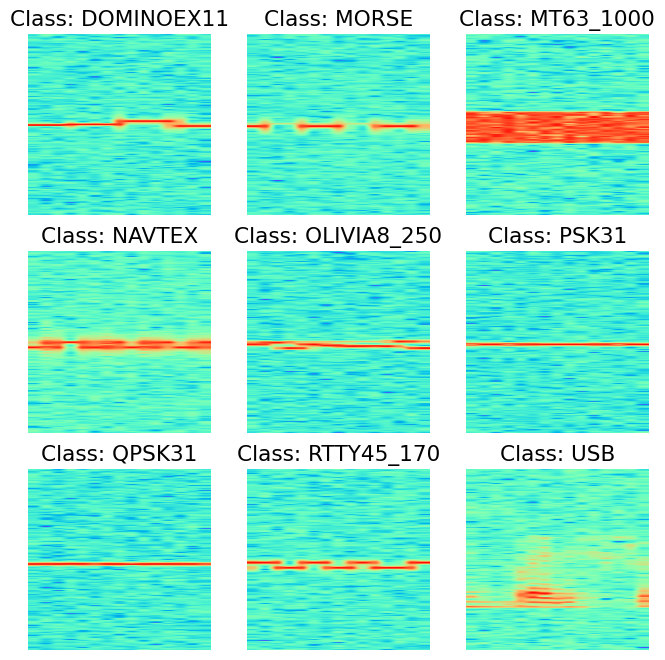}}
    \caption{Spectrogram of a subset of waveforms included in the HF dataset.}
    \label{fig:real_data_fig}
\end{figure}

The data has been created synthetically by first modulating speech, music and text using standard software. Then the signals are cut into short slices. Data is sampled at 6~kHz and consists of 2048 I/Q samples corresponding to 340~msec. The samples are impaired by Gaussian noise, Watterson fading (to account for ionospheric propagation) and random frequency and phase offset. This process is used to generate the data, that is close to real reception signals. The Watterson fading model is a commonly used to model HF communication links.  This channel model is recommended by the International Telecommunication Union (ITU) to capture the effects of frequency spread, frequency offset and differential time delay,  additive white Gaussian noise (AWGN) noise, random frequency offset, and random phase offset, please see \cite{recommendation1994520} for details. Fig.~\ref{fig:real_data_fig} depicts the spectrograms of nine waveform types in the HF dataset. A spectrogram shows the time-frequency representation of a signal, illustrating how the frequency of the signal content evolves over time.

For the classification of waveforms, we use a variant of the Residual Network (ResNet) model. As demonstrated in \cite{Oshea3}, the original ResNet50 architecture did not provide good RF signal classification performance. Instead, a cascade of eight 5-layer residual stacks was proposed and empirically shown to be suitable for signal classification. Since the main contribution of our paper is to present a novel data augmentation methodology that integrates controlled noise in the latent space of the VQ-VAE model, we use the same model architecture from \cite{Oshea3, scholl2019classification} for a fair comparison of results. 
 
	
\begin{table}[tb!]
    \centering
    \caption{Modulation and Baud rate of HF Waveforms \cite{scholl2019classification}.}
    \label{table:mods}
    \small
    \begin{tabular}{ccc}
            \toprule
            Waveform name & Modulation & Baud rate \\
            \midrule
            Morse Code & OOK &	Variable\\		
            PSK31 &	PSK	& 31\\
            PSK63 &	PSK &	63\\
            QPSK31 &	QPSK &	31\\
            RTTY 45/170	& FSK, 170 Hz shift	& 45\\
            RTTY 50/170	& FSK, 170 Hz shift	& 50\\
            RTTY 100/850 &	FSK, 850 Hz shift & 850\\
            Olivia 8/250 &	8-MFSK	& 31\\
            Olivia 16/500 &	16-MFSK	& 31\\
            Olivia 16/1000 &	16-MFSK	& 62\\
            Olivia 32/1000 &	32-MFSK	& 31\\
            DominoEx &	18-MFSK	& 11\\
            MT63 / 1000	& multi-carrier &	10\\
            Navtex / Sitor-B &	FSK, 170 Hz shift &	100\\
            Single-Sideband (upper)	& USB &	-\\
            Single-Sideband (lower)	& LSB &	-\\
            AM broadcast	& AM	&-\\
            HF/radiofax	& Radiofax & -\\ 
            \bottomrule
        \end{tabular}
\end{table}

One of the first steps we took to introduce more variation to the dataset was to apply the data augmentation discussed in \cite{davaslioglu_ssl_2023} to the dataset. This data augmentation consists of five transformations that can be applied randomly to the RF I/Q dataset. These transformations have been applied to self-supervised learning framework where the model is pretrained without the need for any labels \cite{davaslioglu_ssl_2023}. The underlying transformations include adding a DC offset, time-domain shifting (time shift), amplitude scaling (multiplying by a constant factor), zero-masking (nulling a sequence of consecutive samples), and introducing AWGN. DC shift is between 0 and 0.0001, time shift (samples) is between -40 and 40, amplitude scale is between 0.8 and 1.2, zero-masking  (samples) can null up to 25 consecutive samples, and AWGN has a zero mean and a variance of $10^{-5}$.
	
	%
	
\section{Proposed Synthetic Data Generation Methodology}\label{sec:proposed_synth}

Generative models have emerged as powerful tools in machine learning to generate new data samples that resemble a given dataset. These models can be broadly classified into several types, including Variational Autoencoders (VAEs), Generative Adversarial Networks (GANs), and more recent approaches such as Vector Quantized Variational Autoencoders (VQ-VAEs) and diffusion models. VAEs are known for their ability to learn latent representations in a continuous space, enabling smooth interpolation between different data points.  GANs, on the other hand, consist of a generator and a discriminator network, where the generator learns the data distribution and generates synthetic data, while the discriminator evaluates the data as real or fake. However, GANs often face challenges like mode collapse and training instability. VQ-VAEs combine the strengths of VAEs with discrete latent spaces, offering improved performance in generating diverse and multi-class data without the typical pitfalls of GANs. 

Traditional VAEs consist of an encoder network that maps input samples, such as RF waveforms, to a Gaussian distribution by returning the mean $\mu$ and variance $\sigma^2$ of the distribution. In this paper, we use a VQ-VAE that maps input samples to discrete latent variables instead of a continuous distribution. 
VQ-VAEs utilize a Vector Quantizer (VQ) to obtain a discrete latent representation, addressing the issue of posterior collapse—a problem in the standard VAE framework where latents are ignored due to the presence of a powerful autoregressive decoder. By combining the discrete representations with an autoregressive prior, the VQ-VAE model generates high-quality data samples.

The data, $x$, is input to the encoder network which outputs a latent representation, $z_e$. For the latent embedding space, we define $e \in \mathrm{R}^{K\times D}$ where $K$ and $D$ stand for the size of the discrete latent space and dimensionality of each latent vector, $e_i$, respectively. The latent vector $z_e$ is quantized by identifying the nearest neighbor $e_i$ by calculating 
\begin{align}
q(z=k|x)= \begin{cases} 1 & \text{for } k=\arg \min_j \lVert z_e(x)-e_j \rVert_2 \\ 0 & \text{otherwise,} \end{cases}
\end{align}
where each $z_e$ is mapped to the nearest vector embedding in a codebook. We express the quantized latent vector, $z_q$, as
\begin{align}
    z_q(x) = e_k, \quad \text{where } k = \arg\min_j \lVert z_e(x)-e_j \rVert_2. 
\end{align}
The quantized latent vector, $z_q$, is then input into the decoder network to reconstruct the RF signal sample, $\hat{x}$. 

The VQ-VAE is trained using a loss function composed of three key components: the reconstruction loss, the codebook loss, and the commitment loss. The reconstruction loss optimizes the encoder and decoder by measuring the mean squared error (MSE) between the input data $x$ and the reconstructed sample $\hat{x}$. The commitment loss ensures that the latent vectors $z_e$ produced by the encoder remain close to their respective embedding vectors $e_k$. The codebook loss employs Exponential Moving Average (EMA) to adjust the embedding vectors, aligning them more closely with the general trend of the encoder output. We express the loss function as 
\begin{align}
    \ell = \log p(x|z_q(x)) +& \lVert \mathrm{sg}[z_e(x)]- e\rVert_2^2  + \beta \lVert z_e(x) - \mathrm{sg}[e] \rVert_2^2, \label{eq:vqvae_loss}
\end{align}
where $sg[\cdot]$ is the stop-gradient operator, which acts as an identity function during forward computation, but has zero partial derivatives. This effectively constrains its operand to remain a constant that is not updated during backpropagation. In (\ref{eq:vqvae_loss}), the first term is the reconstruction loss. The second term ensures that the embedding vector $e_i$ move towards the encoder outputs $z_e(x)$ using the $\ell_2$ norm. The third term is the commitment loss which encourages the encoder commit to an embedding such that its output does not grow unboundedly. 

\begin{figure}[t!] \centerline{\includegraphics[width=\linewidth]{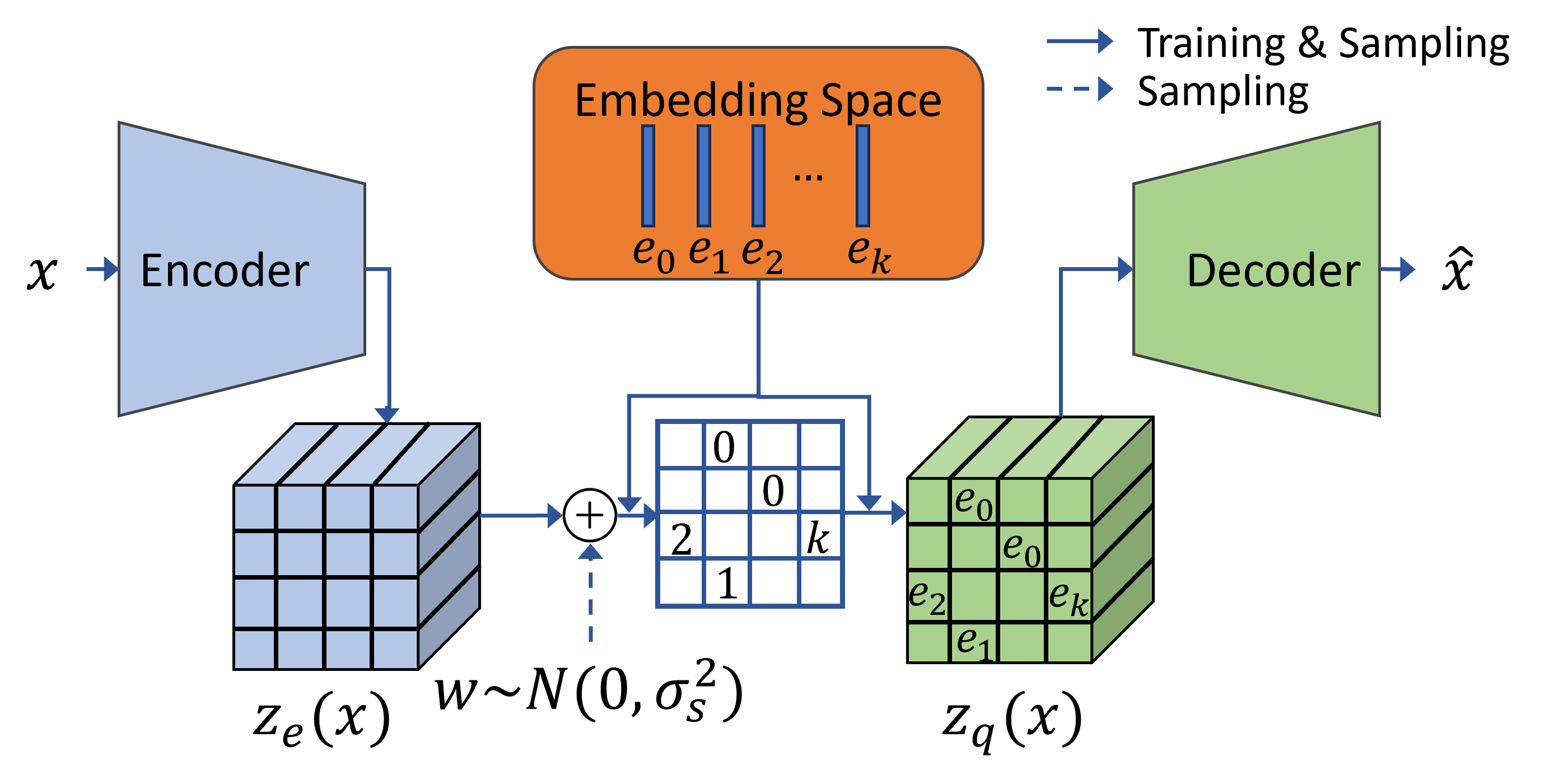}}
    \caption{Proposed VQ-VAE sampling procedure.}
    \label{fig:vqvae_sampling}
\end{figure}

\subsection{VQ-VAE Implementation Details}\label{sec:vqvae_proposed_details}

\begin{table*}[ht!]
    \footnotesize
    \centering
    \caption{Waveform classification performance of different parameters.}
    \label{table:perf_results}
    \resizebox{\textwidth}{!}{%
        \begin{tabular}{cccccccccc}
        \toprule
            Model  &	Learning 	&   Augmentation & 	\multicolumn{2}{c}{VQ-VAE Parameters} 	& Accuracy	& $\Delta$ \% &-10~dB SNR& $\Delta$ \%&25~dB SNR\\
            Name & Rate &  & Perplexity & Noise ($\sigma_s^2$) & &Overall&Accuracy&-10~dB SNR& Accuracy\\
            \midrule
            Base 1	& 0.0001 &	0	& -- &	-- & 	0.9078	& -- &0.6236&--&0.9759\\
            Base 2	& 0.0010 &	0	& --	& -- & 	0.9158	& 0.8\% &0.6361&1.25\%& 0.9825\\
            Base 3	& 0.0010 &	1	& --	&-- &	0.9321 &2.43\% &0.7060&8.24\%&0.9843\\
            Base 4	& 0.0001 &	1	& --	& -- &	0.9337 &2.59\% &0.7180&9.44\%&0.9836\\ \midrule  
            Noise 1	& 0.0001 &	1 &	 0.01 &	0 &	0.9295 &2.17\% &0.7128&8.92\%&0.9834\\
            Noise 2	& 0.0001 &	1 &	0.001 &	0 &	0.9306 &2.28\% &0.6939&7.03\%&0.9855\\
            Noise 3	& 0.0001 &	1 &	0.001 &	0.1 &	0.9371 &2.93\% &0.7282&10.46\%&0.9855\\
            Noise 4	& 0.0001 &	1 &	0.001 &	1 &	0.9397 &3.19\% &0.7508&12.72\%&0.9820\\
            Noise 5	& 0.0001 &	1 &	0.001 &	1+0.1 &	0.9445&3.67\% &0.7638&14.02\%&0.9857\\
            Noise 6	& 0.0001 &	1 &	0.001 &	1.5 &0.9484&4.06\% &0.7822&15.86\%&0.9853\\ 
        \bottomrule
            
        \end{tabular}
    }
\end{table*}

\begin{figure*}[tbh!]    \centering{\includegraphics[width=0.95\textwidth]{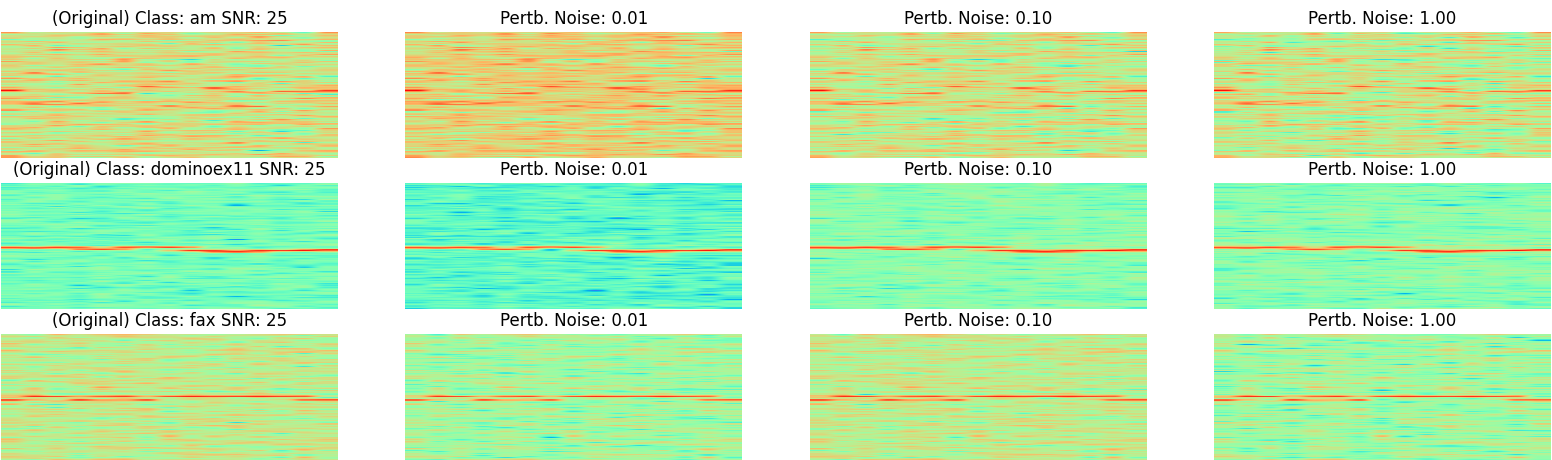}}
    \caption{Spectrogram visualizations of VQ-VAE generated signal samples at varying noise levels.}
    \label{fig:sc2}
\end{figure*}

The solid lines in Fig.~\ref{fig:vqvae_sampling} show the VQ-VAE data processing steps where the input data is encoded, quantized, and decoded. 
There are different approaches to sample synthetic data from a trained VQ-VAE that preserves the label of the samples. These include posterior sampling, sampling from class centers, and interpolation between two points in the latent space. However, we observed that these methods did not provide sufficient diversity for the downstream signal classification model and in most cases, it even decreased our classification accuracy. To address this issue, we have introduced noise in the latent space of the VQ-VAE prior to quantization. After training the VQ-VAE model, noise is added to the signal embedding in the latent space. We represent the noise $w \sim \mathcal{N} \ (0,\sigma_s^2)$ where $\sigma_s^2$ is the latent noise variance. This approach resulted in the generation of diverse data, leading to substantial improvements in classification accuracy. In Section~\ref{sec:performance}, we evaluate the performance of classification models trained on synthetic samples generated using the same VQ-VAE model with different latent space noise variances.

In our implementation, our encoder network uses three 2D convolutional layers that progressively extract and compress features from the input data, followed by a single residual stack. Rectified Linear Unit (ReLU) activation is applied to introduce nonlinearity between the convolutional layers. The final layer of the encoder network employs a pre-quantizer convolutional layer, which reshapes the sample to the embedding dimension.

The VQ layer takes the output of the encoder as its input, reshapes and flattens the input sample, computes the distance between the input vector and each codebook vector, and then maps the input vector to the nearest codebook vector. After mapping, the quantizer updates the codebook using EMA and returns the total loss, embeddings, quantized inputs, and perplexity, which represents the average codebook usage of the input. The perplexity weight determines the rate at which the VQ-VAE model is updated.

The decoder network inverses the operations of the encoder network and reconstructs the original input from the encoded and quantized representations. The decoder network uses a post-quantizer convolutional layer, with its outputs passed through three 2D convolutional transpose layers to progressively upsample and restore the data to its original dimensions.


\section{Performance Evaluation } \label{sec:performance}
In this section, we present our numerical evaluations. We implemented the VQ-VAE model and the classification models in Pytorch using the architectures described in Sections~\ref{sec:system_model} and \ref{sec:vqvae_proposed_details}. We used NVIDIA GeForce RTX 3090 to train and evaluate our results. Each classification model has 31.7~M parameters. Our model achieves an inference time of 276~$\mu$s per sample, with 5.86~giga multiply-accumulate operations (GMACs) per sample, using a batch size of 16.

Table~\ref{table:perf_results} shows the performance of multiple waveform classifiers we trained. We present the overall accuracy of each model as well as those at -10~dB and 25~dB SNR values to evaluate the improvements at the two ends of the SNR range. Base models in Table~\ref{table:perf_results} are trained using only the non-augmented data. When augmentation is 1, the signal transformations discussed in Section~\ref{sec:system_model} are employed. Noise models (Noise~1-6) are trained using the non-augmented data and synthetic data generated with different levels of noise in the latent space. This means that during training of the waveform classifier, reconstructions of the waveform are perturbed in the latent space at different noise levels while maintaining the sample labels.  For Noise~5 model, we pass the data through the VQ-VAE twice with different levels of noise, whereas for all other models, they are input only once. Note that all models presented in Table~\ref{table:perf_results} have the same neural network architecture, initialized using the same weights, and are trained using the same optimizer. We evaluate the effects of different learning rates, augmentations (signal processing transformations), and synthetic data generation with different perplexity weights and latent noise. 

First, we observe that the learning rate and augmentation have significant effect on classification accuracy. Models with augmentation consistently achieve 1.63\% to 2.59\% higher classification accuracy compared to those without augmentation. Models with a learning rate of 0.0001 perform better than those with a learning rate of 0.001, which was consistent for models trained with and without augmentation. The best combination of learning rate and augmentation achieved 93.1\% classification accuracy. 

	
As illustrated in Table.~\ref{table:perf_results}, introducing latent noise to the VQ-VAE during the training phase enhances the overall classification accuracy from 2.93\% to 4.06\% over the Base~1 model across different noise levels. These models are denoted by Noise 1-6. While the benchmark Base~1 model performs well at high SNR values, the benefits of adding latent noise during training are particularly noticeable at lower SNR levels. For example, at an SNR of -10~dB, the Noise~6 model significantly improves classification accuracy from 62.36\% to 78.22\% compared to the Base~1 model. This shows that VQ-VAE is exceptional at introducing significant variety in noisy RF data. The limited increase in classification accuracy at high SNRs ($\geq$10~dB) can be attributed to the diminishing returns phenomenon in deep neural networks 
as they typically approach their performance ceiling at this level. 
Another important result is that while the Noise~6 model improves upon the Base~4 model by approximately 1.47\% across all SNRs, it increases accuracy from 71.80\% to 78.22\% at -10~dB SNR, emphasizing the importance of synthetic data generation beyond signal transformations. 

\begin{figure}[t!]
\centering{\includegraphics[width=0.95\linewidth]{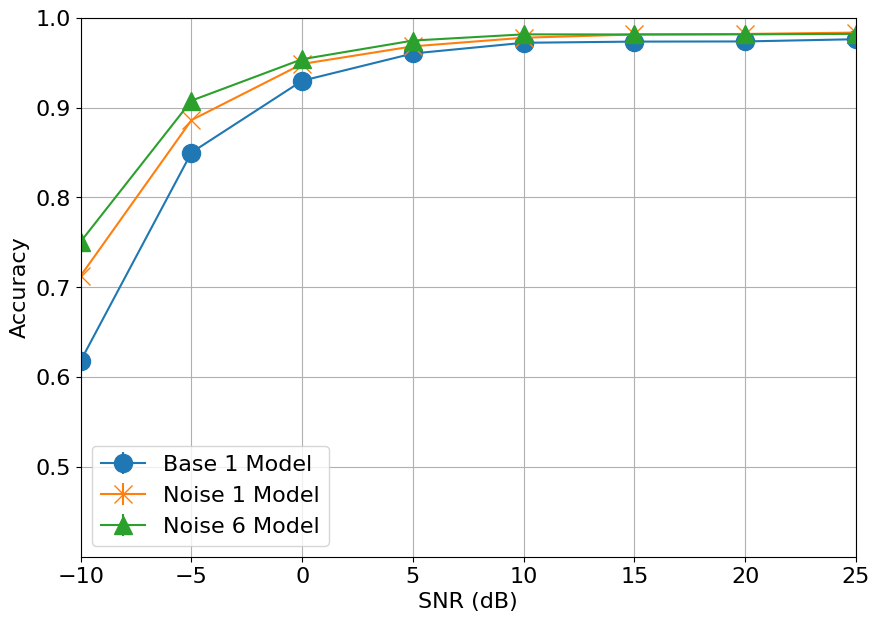}}
\caption{Accuracies of all models across SNRs between -10~dB to 25~dB.}
\label{fig:sc5}
\end{figure}

\begin{figure}[t!]
\centering{\includegraphics[width=0.98\linewidth]{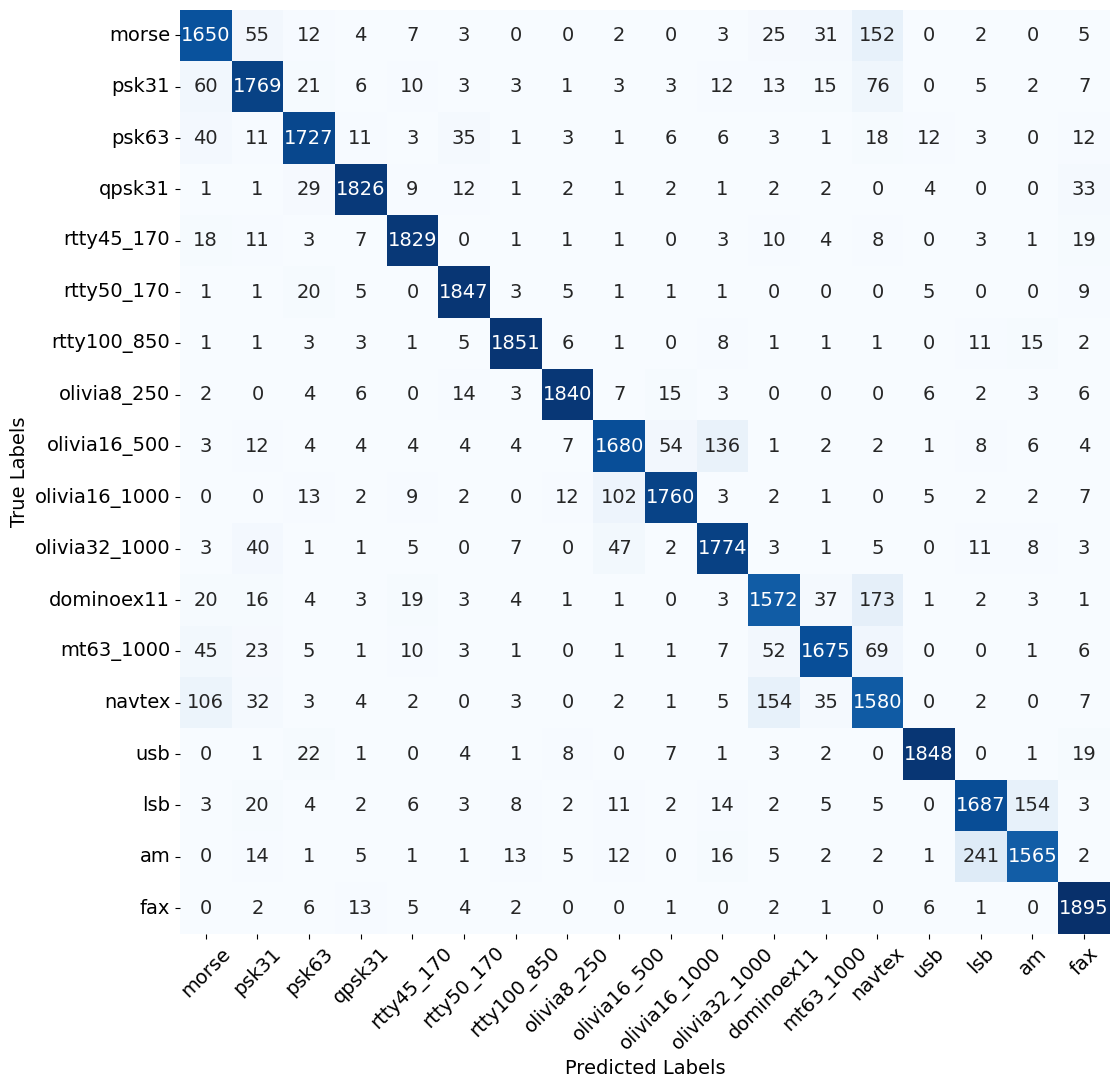}}
\caption{Confusion matrix across all SNRs for Base~1 model.}
\label{fig:CM1}
\end{figure}

\begin{figure}[tbh!]
\centering{\includegraphics[width=0.98\linewidth]{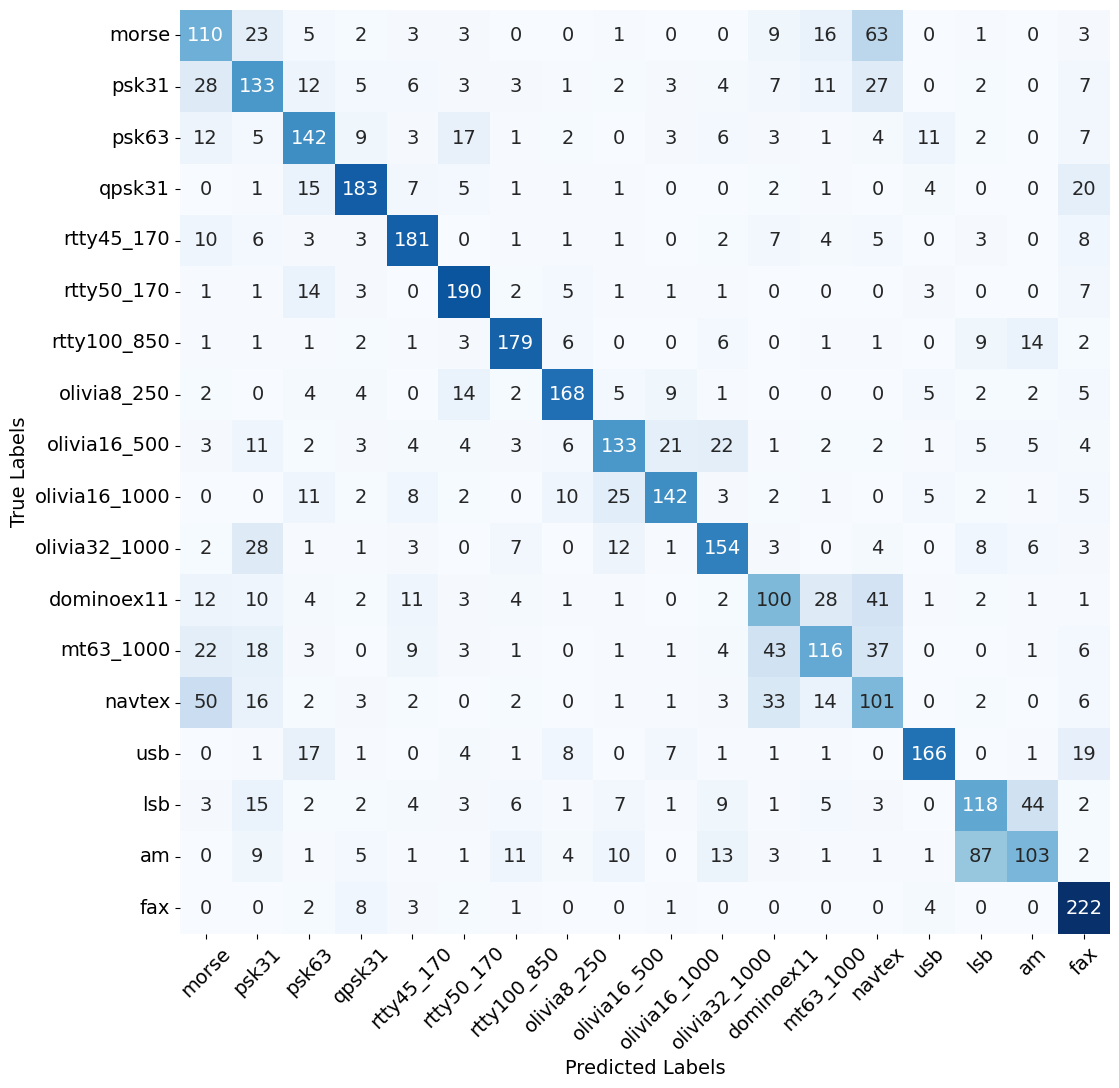}}
\caption{Confusion matrix for -10~dB SNR for Base~1 model.}
\label{fig:CM2}
\end{figure}

\begin{figure}[t!]
\centering{\includegraphics[width=0.98\linewidth]{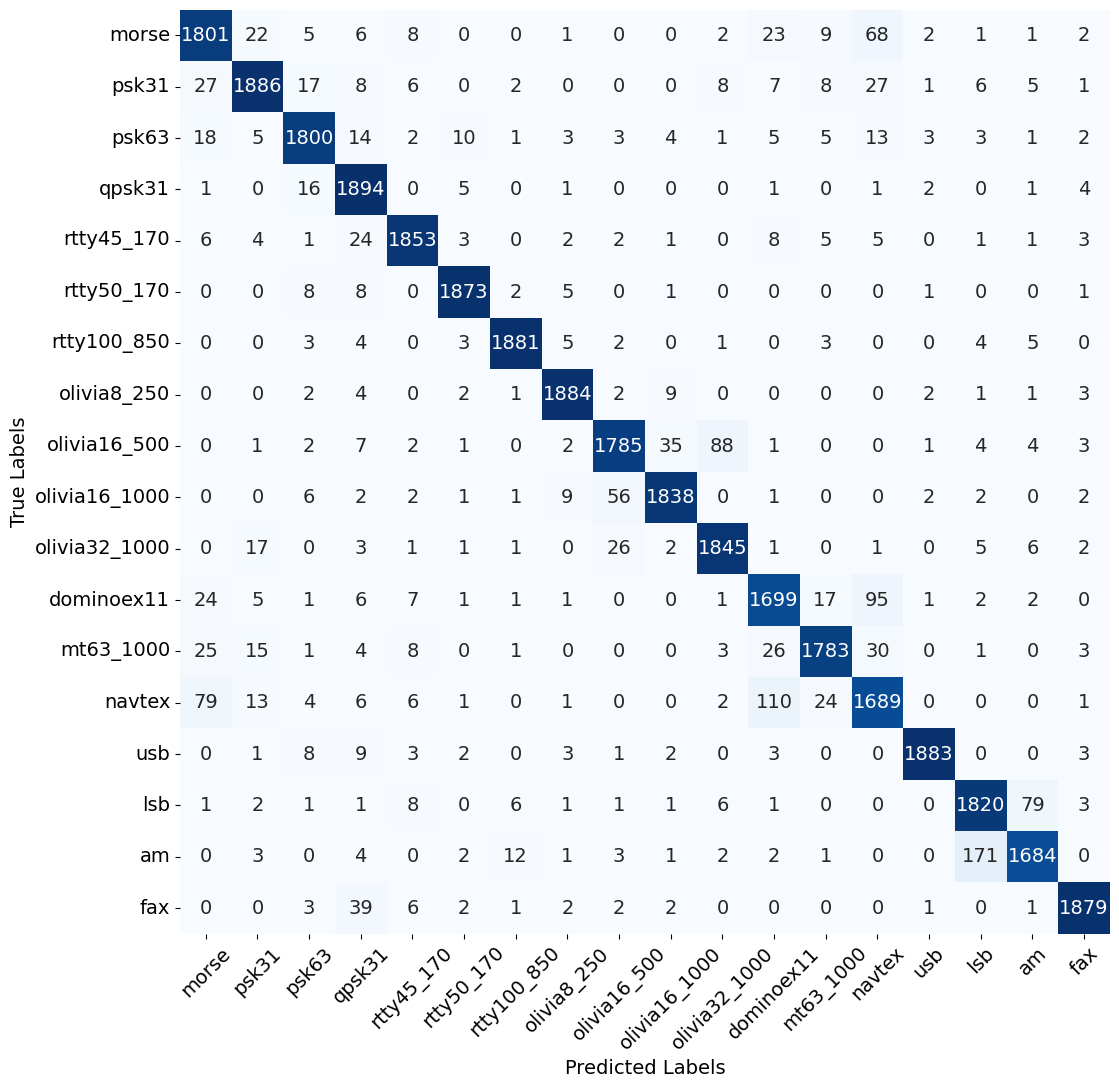}}
\caption{Confusion matrix across all SNRs for Noise~6 model.}
\label{fig:CM3}
\end{figure}

\begin{figure}[t!]
\centering{\includegraphics[width=0.98\linewidth]{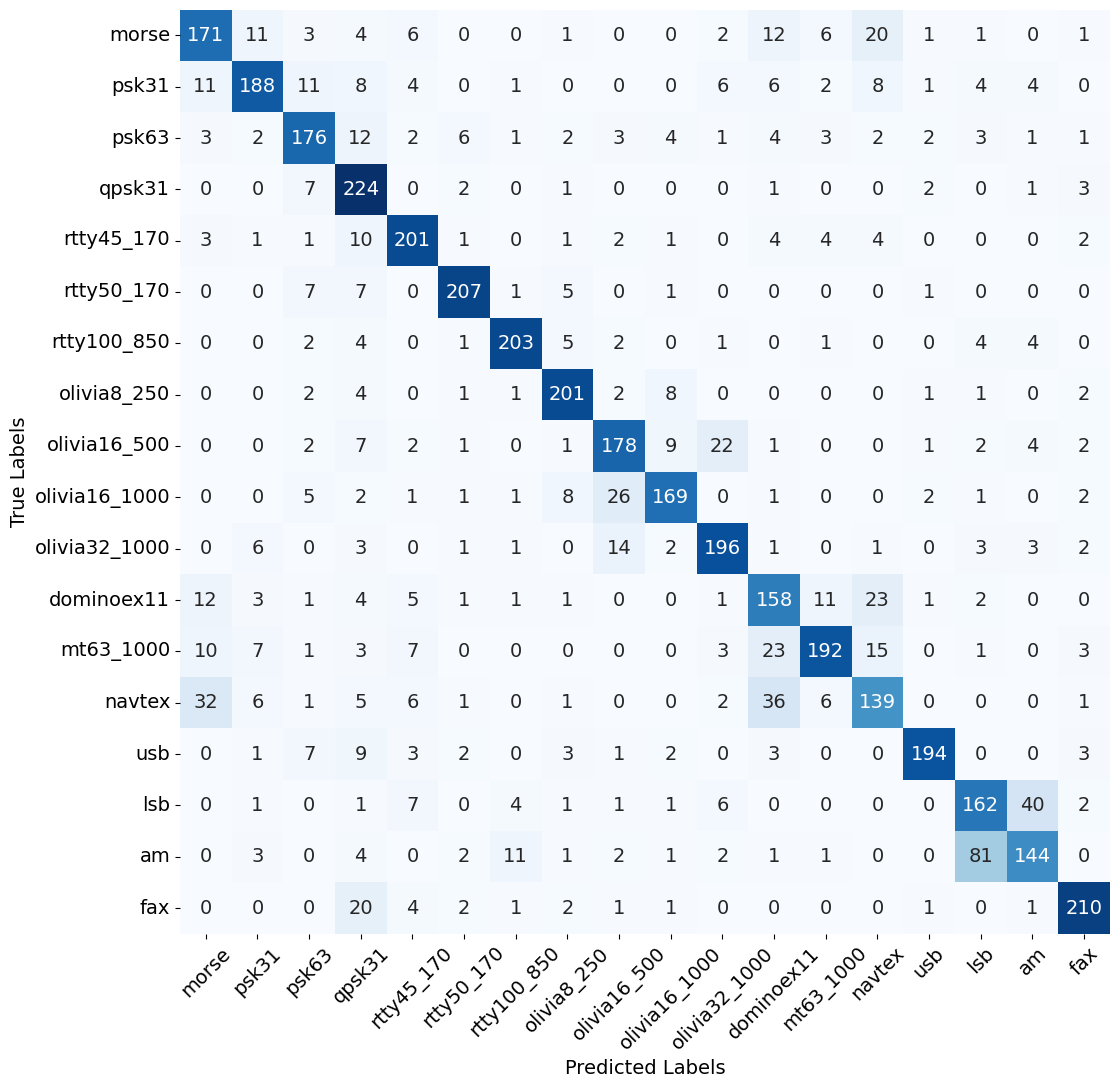}}
\caption{Confusion matrix for -10dB SNR for Noise~6 model.}
\label{fig:CM4}
\end{figure}

Figs.~\ref{fig:CM1} and \ref{fig:CM2} present the confusion matrices across all SNRs and at -10~dB SNR for Base~1 model which is trained without data augmentation, respectively. Similarly, Figs.~\ref{fig:CM3} and \ref{fig:CM4} depict the confusion matrices of Noise~6 model across all SNRs and at -10~dB SNR, respectively. Noise~6 model is trained using the data transformations and synthetic data generated using the VQ-VAE model at $\sigma_s^2=1.5$. Our results show that the confusion matrices for the Base~1 model and the Noise~6 model across all SNRs exhibit similar classification patterns. The primary distinction between these models is their accuracy, where the confusion matrix of the Noise~6 model demonstrates significantly higher classification accuracy than that of the Base~1 model. Both models show slight, yet evident, confusion when classifying between the AM and LSB classes. This is an expected result since the LSB waveform is a form of Single Sideband (SSB) modulation, where only the lower sideband of the AM signal is transmitted, while the carrier and the upper sideband are suppressed to reduce bandwidth and power requirements and increase bandwidth efficiency. Also, despite the inherent similarities among the various Olivia waveforms, which differ by their Multiple Frequency-Shift Keying (MFSK) modulation index and Baud rates, both models are able to differentiate between these classes, underscoring the robustness of these models in accurately classifying waveforms with subtle differences. 

At -10~dB SNR, confusion matrices for both models indicate substantial decline in accuracy. 
due to the distortions noise causes in the spectral characteristics of the signals and makes them harder to distinguish. We observe that both models exhibit errors between the AM and LSB classes, with these errors becoming more pronounced at lower SNR values. 

	
	
\section{Conclusion} \label{sec:Conclusion}

In this paper, we tackle the challenge of RF waveform classification by proposing a methodological approach that utilizes VQ-VAE models to generate synthetic data for augmentation. We have developed and trained a deep neural network capable of classifying 18~different waveform types. Through our experiments, we have evaluated the effects of varying learning rates, signal processing transformations, and data augmentation using synthetic data generated with different perplexity weights and latent noise variances. 
Compared to the baseline model trained solely on real data, our augmented dataset significantly improves classification performance, particularly in low SNR environments. Notably, our best model (Noise~6) has increased the classification accuracy by 4.06\% over the baseline model (Base~1) across all SNRs. At -10~dB SNR, we achieved a substantial 15.86\% improvement in accuracy. The proposed framework provides a robust and systematic approach for spectrum awareness applications and can be used to enhance the reliability of communication in both civilian and tactical applications.
\bibliographystyle{IEEEtran}
\bibliography{refs2_v2}
\end{document}